\documentclass[sigconf, review=false]{acmart}

\usepackage{booktabs}
\usepackage{graphicx}
\setcopyright{rightsretained}

\acmDOI{10.475/123_4}

\acmISBN{123-4567-24-567/08/06}

\acmConference[ICVGIP'24]{15th Indian Conference on Computer Vision, Graphics and Image Processing}{December 2024}{Bangalore, India}
\acmYear{2024}
\copyrightyear{2024}

\acmPrice{15.00}

\begin{document}
\title{Adaptive Multi Scale Document Binarisation Using Vision Mamba}

\author{Mohd. Azfar$^*$}
\affiliation{%
  \institution{Department of Electrical Communication Engineering, Indian Institute of Science}
  \city{Bengaluru}
  \country{India}
}
\email{azfarmohd@iisc.ac.in}

\author{Siddhant Bharadwaj$^*$}
\affiliation{%
  \institution{Department of Electrical Communication Engineering, Indian Institute of Science}
  \city{Bengaluru}
  \country{India}
}
\email{siddhantb@iisc.ac.in}

\author{Ashwin Sasikumar}
\affiliation{%
  \institution{Department of Electronics and Communication, International Institute of Information Technology}
  \city{Bengaluru}
  \country{India}  
}
\email{ashwin.sasikumar@iiitb.ac.in}

\thanks{$^*$Equal contribution}

\renewcommand{\shortauthors}{Azfar, et al.}

\begin{abstract}
Enhancing and preserving the readability of document images, particularly historical ones, is crucial for effective document image analysis. Numerous models have been proposed for this task, including convolutional-based, transformer-based, and hybrid convolutional-transformer architectures. While hybrid models address the limitations of purely convolutional or transformer-based methods, they often suffer from issues like quadratic time complexity. In this work, we propose a Mamba-based architecture for document binarisation, which efficiently handles long sequences by scaling linearly and optimizing memory usage. Additionally, we introduce novel modifications to the skip connections by incorporating Difference of Gaussians (DoG) features, inspired by conventional signal processing techniques. These multiscale high-frequency features enable the model to produce high-quality, detailed outputs.
\end{abstract}



\maketitle

\section{Introduction and Related Work}
Document image binarisation is a critical task in various document processing applications. Several approaches leveraging deep learning have emerged over the years. Early methods focused on adaptive thresholding techniques such as the work by Bradley and Roth \cite{bradley2007adaptive}, which utilized integral images for thresholding. However, with advancements in deep learning, the focus shifted towards using models like LSTM for sequence learning in binarisation, as proposed by Afzal et al. \cite{afzal2015document}. This approach treats binarisation as a sequence classification problem and processes image sequences with a pixel-level model. Other methods, such as the scalable fully convolutional network (FCN) presented by Bezmaternykh et al. \cite{bezmaternykh2019unet}, aim to provide more efficient solutions through convolutional architectures, targeting larger document sets for binarisation.

\section{Method}
Given a document image $X \in \mathbb{R}^{H \times W \times C}$, where $H$, $W$, and $C$ are the height, width, and channels of the image respectively, our objective is to predict the corresponding binarised image $Y \in \mathbb{R}^{H \times W}$. We propose \textbf{AMSDB-VMamba}, a U-Net-based architecture consisting of: 1) a Mamba-based encoder\cite{gu2023mamba}, pre-trained on the large-scale ImageNet dataset, to capture relevant features, 2) a decoder with multiple up-sampling blocks that capture semantics at different depths to predict the binarised image, and 3) multiscale high-frequency feature extraction at various encoder depths using Difference of Gaussian (DoG) function.

\subsection{Encoder and Decoder}
Mamba-based architectures, known for their higher accuracy, lower computational burden, and reduced memory consumption in natural language processing, are not directly applicable to images due to limited receptive fields, but this can be addressed by using the Mamba block as the encoder's basic unit\cite{liu2024vmamba}, which expands a 2D input feature map \( z \) along four directions \( v \in \{1, 2, 3, 4\} \) as \( z_v = \text{expand}(z, v) \), processes each sequence via a state-space model \( \bar{z}_v = S6(z_v) \), and merges the outputs as \( \bar{z} = \text{merge}(\bar{z}_1, \bar{z}_2, \bar{z}_3, \bar{z}_4) \), effectively capturing spatial dependencies in 2D images.
\newline
The native up-sample block in the U-shaped architecture is modified to include an additional convolutional block with a residual connection and a segmentation head at each scale for deep supervision.

\begin{figure}[h]
    \centering
    \includegraphics[width=0.3\textwidth]{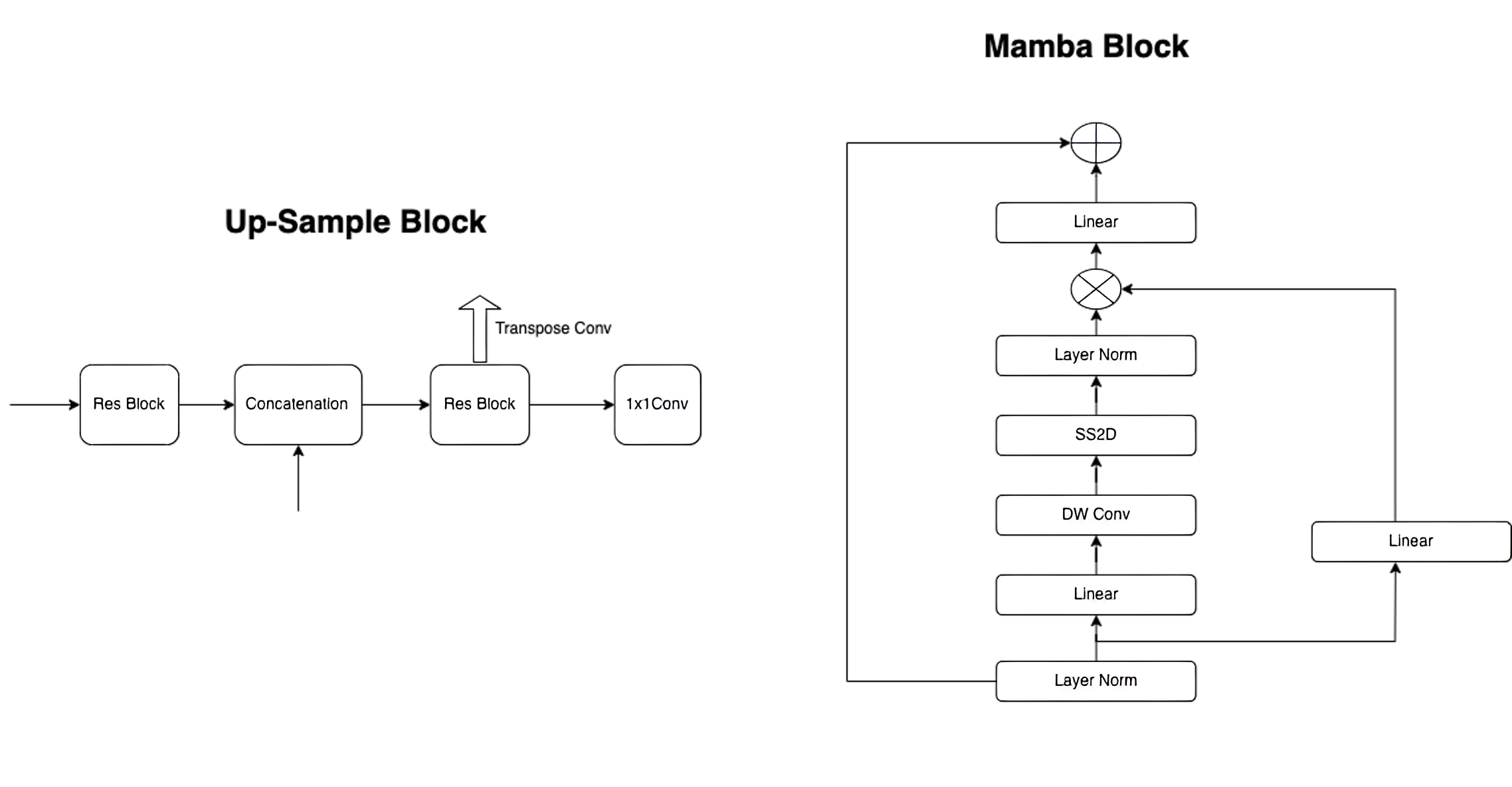}
    \caption{Schematic of Mamba block \& Upsample block}
    \label{fig:blocks}  
\end{figure}

\subsection{Skip Connections and the Role of DoG Features in U-Net}
Skip connections are a fundamental component of the U-Net architecture, bridge the gap between low-level details and high-level semantics. They facilitate the direct transfer of feature maps from early layers of the encoder to corresponding layers in the decoder. This mechanism is crucial for preserving spatial information that might be lost during the downsampling process, ensuring that fine details—such as edges and textures—are retained in the final output.

We propose using the Difference of Gaussians (DoG) function over the low-level semantics of the encoder layers instead of directly using skip connections as in \cite{liu2024swin}, where DoG is mathematically defined as \( \text{DoG}(X ,\sigma_{1} ,\sigma_{2}) = G(X; \sigma_{1}) - G(X; \sigma_{2}) \), with \( G(X; \sigma) \) evaluated for all pixels of a feature map using a Gaussian kernel \( K(x, y) = \frac{1}{2\pi \sigma^2} \exp\left(-\frac{(x - i)^2 + (y - j)^2}{2\sigma^2}\right) \), where \( (x, y) \) are coordinates relative to the center pixel \( (i,j) \), \( \sigma \) is the standard deviation, and \( k \times k \) is the kernel size, effectively capturing high-frequency components from the encoder layers to extract fine-grained textures crucial for producing high-quality detailed outputs.
\newline
In the context of U-Net, incorporating multiscale DoG features by using different scales of sigma values helps extract multiresolution features from the encoder, and we introduce learnable weights \( w_i \) associated with each scale \( i \), shared across the layers at a particular depth, which are learned using the autograd package, and mathematically defined as \( F_{\text{DoG}}(X) = \sum_{i=1}^{N} w_i \cdot \text{DoG}(X; \sigma_{1,i}, \sigma_{2,i}) \), where \( \sigma_{1,i} = \sigma_{0} \cdot 2^{\frac{i}{N}} \) and \( \sigma_{2,i} = \sigma_{1,i} \cdot 2^{\frac{1}{N}} \), allowing the network to adaptively prioritize frequency components and optimize reconstruction quality.The integration of skip connections and multiscale DoG features not only preserves essential spatial details but also enhances the model’s ability to adapt to various imaging challenges. This combination ultimately leads to improved performance in dense prediction tasks, such as image segmentation and reconstruction.

\section{Implementation Details}
During training, we split each degraded image and the corresponding binarised ground truth into overlapping patches of size $(128 \times 128)$, which are used for training and validation. For training, we used augmentations such as random cropping, flipping, and horizontal and vertical rotation. The batch size used for training is 32. We evaluate our model's performance on three metrics: Peak Signal-to-Noise Ratio (PSNR), F-Measure (FM), and pseudo-F-measure ($F_{\text{ps}}$).

We train our model using the leave-one-out approach, which is a standard process in papers addressing document binarisation. In the leave-one-out approach, we consider images of one particular year as a testing dataset, and the remaining data from all the other years is used for training. The model is trained on DIBCO 2009 \cite{gatos2009icdar}, DIBCO 2010 \cite{pratikakis2010hdibco}, DIBCO 2011 \cite{pratikakis2011icdar}, DIBCO 2012 \cite{pratikakis2012icfhr}, DIBCO 2013 \cite{pratikakis2013icdar}, DIBCO 2014 \cite{ntirogiannis2014icfhr}, and test on DIBCO 2016 \cite{pratikakis2016icfhr}.

\begin{figure}[h]
    \centering
    \includegraphics[width=0.5\textwidth]{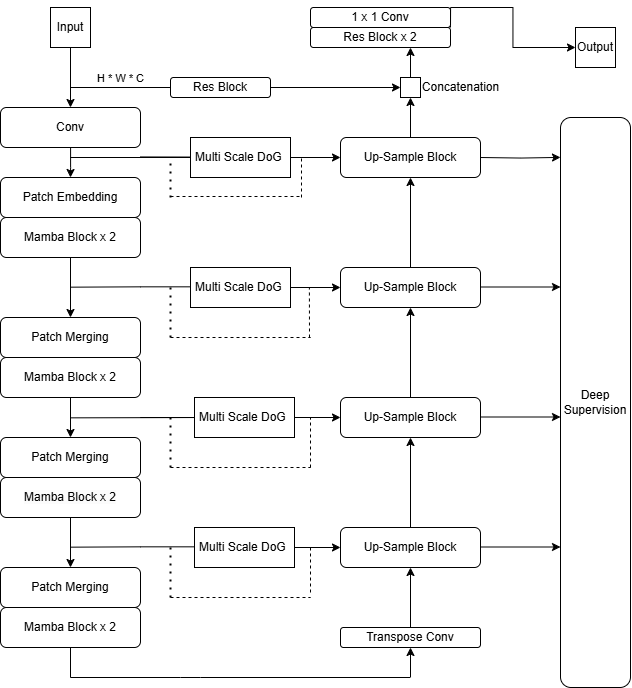}
    \caption{Overview of AMSDB-VMamba for document image binarisation}
    \label{fig:overview}
\end{figure}

\begin{figure}[h]
    \centering
    \includegraphics[width=0.5\textwidth]{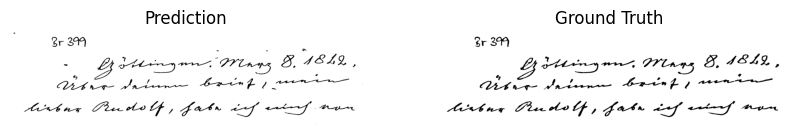}
    \caption{Comparison of our model's prediction with ground truth}
    \label{fig:comparison}  
\end{figure}

\begin{table}[h]
\centering
\caption{Comparative results of the proposed method and other state-of-the-art methods on the DIBCO 2016 dataset.}
\label{tab:comparative}  
\footnotesize
\begin{tabular}{@{}lccc@{}}
\toprule
Methods & PSNR $\uparrow$ & FM $\uparrow$ & Fps $\uparrow$ \\
\midrule
Otsu \cite{otsu1979threshold} & 17.82 & 86.74 & 88.86 \\
Sauvola et al. \cite{sauvola2000adaptive} & 17.15 & 83.68 & 87.48 \\
Tensemeyer et al. \cite{tensmeyer2017document} & 18.67 & 89.52 & 93.76 \\
Howe et al. \cite{howe2011laplacian} & 18.05 & 87.47 & 92.28 \\
\midrule
AMSDB-VMamba (Ours) & 17.79 & 81.36 & 84.89 \\
AMSDB-VMamba (With DoG Connections) (Ours) & 18.06 & 88.13 & 92.87 \\
AMSDB-VMamba (With DoG-Residual Connections) (Ours) & \textbf{18.27} & \textbf{88.42} & \textbf{93.10} \\
\bottomrule
\end{tabular}
\end{table}

\begin{table}[h]
\centering
\caption{Ablation of the proposed method and comparison with the state-of-the-art on the DIBCO 2016 dataset.}
\label{tab:ablation}  
\scriptsize
\begin{tabular}{@{}lcccccc@{}}
\toprule
Methods & PSNR $\uparrow$ & FM $\uparrow$ & Fps $\uparrow$ & Params & Milliseconds/Image \\
\midrule
TransDocUNet \cite{biswas2024transdocunet} & 18.69 & 89.90 & 93.88 & 105M & 190 \\
\midrule
AMSDB-VMamba (Ours) & 17.79 & 81.36 & 84.89 & 59M & 80 \\
AMSD-VMamba (With DoG) (Ours) & 18.06 & 88.13 & 92.87 & 59M & 90 \\
AMSD-VMamba (DoG + Residual) (Ours) & \textbf{18.27} & \textbf{88.42} & \textbf{93.10} & 59M & 90 \\
\bottomrule
\end{tabular}
\end{table}

\section{Conclusion}
Our proposed model AMSD-VMamba, as tested on DIBCO 2016, shows performance comparable to state-of-the-art methods for document binarisation despite requiring significantly fewer parameters; even at inference time, the model is faster than the state-of-the-art TransDocUNet. This architecture has the potential for further exploration in developing fast segmentation models, particularly in applications that demand intricate details, such as medical image segmentation and real-time video processing.
\newpage
\bibliography{references}


\begin{thebibliography}{18}


\ifx \showCODEN    \undefined \def \showCODEN     #1{\unskip}     \fi
\ifx \showDOI      \undefined \def \showDOI       #1{#1}\fi
\ifx \showISBNx    \undefined \def \showISBNx     #1{\unskip}     \fi
\ifx \showISBNxiii \undefined \def \showISBNxiii  #1{\unskip}     \fi
\ifx \showISSN     \undefined \def \showISSN      #1{\unskip}     \fi
\ifx \showLCCN     \undefined \def \showLCCN      #1{\unskip}     \fi
\ifx \shownote     \undefined \def \shownote      #1{#1}          \fi
\ifx \showarticletitle \undefined \def \showarticletitle #1{#1}   \fi
\ifx \showURL      \undefined \def \showURL       {\relax}        \fi
\providecommand\bibfield[2]{#2}
\providecommand\bibinfo[2]{#2}
\providecommand\natexlab[1]{#1}
\providecommand\showeprint[2][]{arXiv:#2}

\bibitem[\protect\citeauthoryear{Afzal, Pastor-Pellicer, Shafait, Breuel, Dengel, and Liwicki}{Afzal et~al\mbox{.}}{2015}]%
        {afzal2015document}
\bibfield{author}{\bibinfo{person}{Muhammad~Zeshan Afzal}, \bibinfo{person}{Joan Pastor-Pellicer}, \bibinfo{person}{Faisal Shafait}, \bibinfo{person}{Thomas~M Breuel}, \bibinfo{person}{Andreas Dengel}, {and} \bibinfo{person}{Marcus Liwicki}.} \bibinfo{year}{2015}\natexlab{}.
\newblock \showarticletitle{Document Image Binarization Using LSTM: A Sequence Learning Approach}. In \bibinfo{booktitle}{\emph{Proceedings of the 3rd International Workshop on Historical Document Imaging and Processing}}. \bibinfo{pages}{79--84}.
\newblock


\bibitem[\protect\citeauthoryear{Bezmaternykh, Ilin, and Nikolaev}{Bezmaternykh et~al\mbox{.}}{2019}]%
        {bezmaternykh2019unet}
\bibfield{author}{\bibinfo{person}{Pavel~Vladimirovich Bezmaternykh}, \bibinfo{person}{Dmitrii~Alexeevich Ilin}, {and} \bibinfo{person}{Dmitry~Petrovich Nikolaev}.} \bibinfo{year}{2019}\natexlab{}.
\newblock \showarticletitle{U-Net-bin: Hacking the Document Image Binarization Contest}.
\newblock \bibinfo{journal}{\emph{Pattern Recognition and Image Analysis}} \bibinfo{volume}{43}, \bibinfo{number}{5} (\bibinfo{year}{2019}), \bibinfo{pages}{825--832}.
\newblock


\bibitem[\protect\citeauthoryear{Biswas, Sarkhel, Roy, and Pal}{Biswas et~al\mbox{.}}{2024}]%
        {biswas2024transdocunet}
\bibfield{author}{\bibinfo{person}{Risab Biswas}, \bibinfo{person}{Soumik Sarkhel}, \bibinfo{person}{Swalpa~Kumar Roy}, {and} \bibinfo{person}{Umapada Pal}.} \bibinfo{year}{2024}\natexlab{}.
\newblock \showarticletitle{TransDocUNet: A Transformer-Based UNet Architecture for Degraded Document Image Binarization}. In \bibinfo{booktitle}{\emph{Proceedings of the Fourteenth Indian Conference on Computer Vision, Graphics and Image Processing}} \emph{(\bibinfo{series}{ICVGIP '23})}. \bibinfo{publisher}{Association for Computing Machinery}, \bibinfo{pages}{1--9}.
\newblock


\bibitem[\protect\citeauthoryear{Bradley and Roth}{Bradley and Roth}{2007}]%
        {bradley2007adaptive}
\bibfield{author}{\bibinfo{person}{Derek Bradley} {and} \bibinfo{person}{Gerhard Roth}.} \bibinfo{year}{2007}\natexlab{}.
\newblock \showarticletitle{Adaptive Thresholding Using the Integral Image}.
\newblock \bibinfo{journal}{\emph{Journal of Graphics Tools}} \bibinfo{volume}{12}, \bibinfo{number}{2} (\bibinfo{year}{2007}), \bibinfo{pages}{13--21}.
\newblock


\bibitem[\protect\citeauthoryear{Gatos, Ntirogiannis, and Pratikakis}{Gatos et~al\mbox{.}}{2009}]%
        {gatos2009icdar}
\bibfield{author}{\bibinfo{person}{Basilis Gatos}, \bibinfo{person}{Konstantinos Ntirogiannis}, {and} \bibinfo{person}{Ioannis Pratikakis}.} \bibinfo{year}{2009}\natexlab{}.
\newblock \showarticletitle{ICDAR 2009 Document Image Binarization Contest (DIBCO 2009)}. In \bibinfo{booktitle}{\emph{2009 10th International Conference on Document Analysis and Recognition}}. IEEE, \bibinfo{pages}{1375--1382}.
\newblock


\bibitem[\protect\citeauthoryear{Gu and Dao}{Gu and Dao}{2023}]%
        {gu2023mamba}
\bibfield{author}{\bibinfo{person}{Albert Gu} {and} \bibinfo{person}{Tri Dao}.} \bibinfo{year}{2023}\natexlab{}.
\newblock \showarticletitle{Mamba: Linear-Time Sequence Modeling with Selective State Spaces}.
\newblock \bibinfo{journal}{\emph{arXiv preprint arXiv:2312.00752}} (\bibinfo{year}{2023}).
\newblock


\bibitem[\protect\citeauthoryear{Howe}{Howe}{2011}]%
        {howe2011laplacian}
\bibfield{author}{\bibinfo{person}{Nicholas~R Howe}.} \bibinfo{year}{2011}\natexlab{}.
\newblock \showarticletitle{A Laplacian Energy for Document Binarization}. In \bibinfo{booktitle}{\emph{2011 International Conference on Document Analysis and Recognition}}. IEEE, \bibinfo{pages}{6--10}.
\newblock


\bibitem[\protect\citeauthoryear{Liu, Yang, Zhou, Xi, Yu, Li, Liang, Shi, Yu, Zhang, and Zheng}{Liu et~al\mbox{.}}{2024b}]%
        {liu2024swin}
\bibfield{author}{\bibinfo{person}{Jiacheng Liu}, \bibinfo{person}{Hongyi Yang}, \bibinfo{person}{Hong-Yu Zhou}, \bibinfo{person}{Yixuan Xi}, \bibinfo{person}{Lixuan Yu}, \bibinfo{person}{Chen Li}, \bibinfo{person}{Yongkang Liang}, \bibinfo{person}{Guangming Shi}, \bibinfo{person}{Yizhou Yu}, \bibinfo{person}{Shanshan Zhang}, {and} \bibinfo{person}{Huazhu Zheng}.} \bibinfo{year}{2024}\natexlab{b}.
\newblock \showarticletitle{Swin-UMamba: Mamba-Based UNet with ImageNet-Based Pretraining}. In \bibinfo{booktitle}{\emph{International Conference on Medical Image Computing and Computer-Assisted Intervention}}. \bibinfo{publisher}{Springer Nature Switzerland}, \bibinfo{pages}{615--625}.
\newblock


\bibitem[\protect\citeauthoryear{Liu, Tian, Zhao, Yu, Xie, Wang, Ye, and Liu}{Liu et~al\mbox{.}}{2024a}]%
        {liu2024vmamba}
\bibfield{author}{\bibinfo{person}{Yue Liu}, \bibinfo{person}{Yunjie Tian}, \bibinfo{person}{Yuzhong Zhao}, \bibinfo{person}{Hongtian Yu}, \bibinfo{person}{Lingxi Xie}, \bibinfo{person}{Yaowei Wang}, \bibinfo{person}{Qixiang Ye}, {and} \bibinfo{person}{Yunfan Liu}.} \bibinfo{year}{2024}\natexlab{a}.
\newblock \showarticletitle{Vmamba: Visual State Space Model}.
\newblock \bibinfo{journal}{\emph{arXiv preprint arXiv:2401.10166}} (\bibinfo{year}{2024}).
\newblock


\bibitem[\protect\citeauthoryear{Ntirogiannis, Gatos, and Pratikakis}{Ntirogiannis et~al\mbox{.}}{2014}]%
        {ntirogiannis2014icfhr}
\bibfield{author}{\bibinfo{person}{Konstantinos Ntirogiannis}, \bibinfo{person}{Basilis Gatos}, {and} \bibinfo{person}{Ioannis Pratikakis}.} \bibinfo{year}{2014}\natexlab{}.
\newblock \showarticletitle{ICFHR2014 Competition on Handwritten Document Image Binarization (H-DIBCO 2014)}. In \bibinfo{booktitle}{\emph{2014 14th International Conference on Frontiers in Handwriting Recognition}}. IEEE, \bibinfo{pages}{809--813}.
\newblock


\bibitem[\protect\citeauthoryear{Otsu}{Otsu}{1979}]%
        {otsu1979threshold}
\bibfield{author}{\bibinfo{person}{Nobuyuki Otsu}.} \bibinfo{year}{1979}\natexlab{}.
\newblock \showarticletitle{A Threshold Selection Method from Gray-Level Histograms}.
\newblock \bibinfo{journal}{\emph{IEEE Transactions on Systems, Man, and Cybernetics}} \bibinfo{volume}{9}, \bibinfo{number}{1} (\bibinfo{year}{1979}), \bibinfo{pages}{62--66}.
\newblock


\bibitem[\protect\citeauthoryear{Pratikakis, Gatos, and Ntirogiannis}{Pratikakis et~al\mbox{.}}{2010}]%
        {pratikakis2010hdibco}
\bibfield{author}{\bibinfo{person}{Ioannis Pratikakis}, \bibinfo{person}{Basilis Gatos}, {and} \bibinfo{person}{Konstantinos Ntirogiannis}.} \bibinfo{year}{2010}\natexlab{}.
\newblock \showarticletitle{H-DIBCO 2010 - Handwritten Document Image Binarization Competition}. In \bibinfo{booktitle}{\emph{2010 12th International Conference on Frontiers in Handwriting Recognition}}. IEEE, \bibinfo{pages}{727--732}.
\newblock


\bibitem[\protect\citeauthoryear{Pratikakis, Gatos, and Ntirogiannis}{Pratikakis et~al\mbox{.}}{2011}]%
        {pratikakis2011icdar}
\bibfield{author}{\bibinfo{person}{Ioannis Pratikakis}, \bibinfo{person}{Basilis Gatos}, {and} \bibinfo{person}{Konstantinos Ntirogiannis}.} \bibinfo{year}{2011}\natexlab{}.
\newblock \showarticletitle{ICDAR 2011 Document Image Binarization Contest (DIBCO 2011)}. In \bibinfo{booktitle}{\emph{2011 International Conference on Document Analysis and Recognition}}. IEEE, \bibinfo{pages}{1506--1510}.
\newblock


\bibitem[\protect\citeauthoryear{Pratikakis, Gatos, and Ntirogiannis}{Pratikakis et~al\mbox{.}}{2012}]%
        {pratikakis2012icfhr}
\bibfield{author}{\bibinfo{person}{Ioannis Pratikakis}, \bibinfo{person}{Basilis Gatos}, {and} \bibinfo{person}{Konstantinos Ntirogiannis}.} \bibinfo{year}{2012}\natexlab{}.
\newblock \showarticletitle{ICFHR 2012 Competition on Handwritten Document Image Binarization (H-DIBCO 2012)}. In \bibinfo{booktitle}{\emph{2012 International Conference on Frontiers in Handwriting Recognition}}. IEEE, \bibinfo{pages}{817--822}.
\newblock


\bibitem[\protect\citeauthoryear{Pratikakis, Gatos, and Ntirogiannis}{Pratikakis et~al\mbox{.}}{2013}]%
        {pratikakis2013icdar}
\bibfield{author}{\bibinfo{person}{Ioannis Pratikakis}, \bibinfo{person}{Basilis Gatos}, {and} \bibinfo{person}{Konstantinos Ntirogiannis}.} \bibinfo{year}{2013}\natexlab{}.
\newblock \showarticletitle{ICDAR 2013 Document Image Binarization Contest (DIBCO 2013)}. In \bibinfo{booktitle}{\emph{2013 12th International Conference on Document Analysis and Recognition}}. IEEE, \bibinfo{pages}{1471--1476}.
\newblock


\bibitem[\protect\citeauthoryear{Pratikakis, Zagoris, Barlas, and Gatos}{Pratikakis et~al\mbox{.}}{2016}]%
        {pratikakis2016icfhr}
\bibfield{author}{\bibinfo{person}{Ioannis Pratikakis}, \bibinfo{person}{Konstantinos Zagoris}, \bibinfo{person}{George Barlas}, {and} \bibinfo{person}{Basilis Gatos}.} \bibinfo{year}{2016}\natexlab{}.
\newblock \showarticletitle{ICFHR2016 Handwritten Document Image Binarization Contest (H-DIBCO 2016)}. In \bibinfo{booktitle}{\emph{2016 15th International Conference on Frontiers in Handwriting Recognition (ICFHR)}}. IEEE, \bibinfo{pages}{619--623}.
\newblock


\bibitem[\protect\citeauthoryear{Sauvola and Pietik{\"a}inen}{Sauvola and Pietik{\"a}inen}{2000}]%
        {sauvola2000adaptive}
\bibfield{author}{\bibinfo{person}{Jaakko Sauvola} {and} \bibinfo{person}{Matti Pietik{\"a}inen}.} \bibinfo{year}{2000}\natexlab{}.
\newblock \showarticletitle{Adaptive Document Image Binarization}.
\newblock \bibinfo{journal}{\emph{Pattern Recognition}} \bibinfo{volume}{33}, \bibinfo{number}{2} (\bibinfo{year}{2000}), \bibinfo{pages}{225--236}.
\newblock


\bibitem[\protect\citeauthoryear{Tensmeyer and Martinez}{Tensmeyer and Martinez}{2017}]%
        {tensmeyer2017document}
\bibfield{author}{\bibinfo{person}{Chris Tensmeyer} {and} \bibinfo{person}{Tony Martinez}.} \bibinfo{year}{2017}\natexlab{}.
\newblock \showarticletitle{Document Image Binarization with Fully Convolutional Neural Networks}. In \bibinfo{booktitle}{\emph{2017 14th IAPR International Conference on Document Analysis and Recognition (ICDAR)}}, Vol.~\bibinfo{volume}{1}. IEEE, \bibinfo{pages}{99--104}.
\newblock


\end{thebibliography}

\end{document}